\documentclass{isprs} % isprs class modified 23-04-2019 (Dennis Wittich)
\usepackage{setspace}
\usepackage{geometry} % added 27-02-2014 Markus Englich
\usepackage{epstopdf}
\usepackage[labelsep=period]{caption}  % added 14-04-2016 Markus Englich - Recommendation by Sebastian Brocks
\usepackage{amsmath}
\usepackage[british]{babel} 
\usepackage[hang]{footmisc}
\usepackage[authoryear]{natbib}
\usepackage{placeins}
\usepackage{here}
\usepackage{color}
\usepackage{subfig}

\begin{document}

\title{Semi-Supervised Segmentation of Concrete Aggregate Using Consensus Regularisation and Prior Guidance}

\author{
Max Coenen\textsuperscript{1,}\thanks{Corresponding author} , Tobias Schack\textsuperscript{1}, Dries Beyer\textsuperscript{1}, Christian Heipke\textsuperscript{2}, Michael Haist\textsuperscript{1}
}

\address{
	\textsuperscript{1 } Institute of Building Materials Science, Leibniz University Hannover, Germany\\
	(m.coenen, t.schack, d.beyer, haist)@baustoff.uni-hannover.de\\
	\textsuperscript{2 } Institute of Photogrammetry and GeoInformation, Leibniz University Hannover, Germany\\
	heipke@ipi.uni-hannover.de
	%\textsuperscript{3 }third affiliation\\
}

% If the corresponding author is NOT the final author, always add a % space before the subsequent comma, i.e.
% first author name\textsuperscript{a,}\thanks{Corresponding author} , % second author name \textsuperscript{b}, etc.
% thanks to Niclas Borlin 05-05-2016

\icwg{}   %This field is optional.

% KAO: Use times symbol
\abstract{
In order to leverage and profit from unlabelled data, semi-supervised frameworks for semantic segmentation based on consistency training have been proven to be powerful tools to significantly improve the performance of purely supervised segmentation learning.
However, the consensus principle behind consistency training has at least one drawback, which we identify in this paper: imbalanced label distributions within the data. 
To overcome the limitations of standard consistency training, we propose a novel semi-supervised framework for semantic segmentation, introducing additional losses based on prior knowledge.  
Specifically, we propose a light-weight architecture consisting of a shared encoder and a main decoder, which is trained in a supervised manner. An auxiliary decoder is added as additional branch in order to make use of unlabelled data based on consensus training, and we add additional constraints derived from prior information on the class distribution and on auto-encoder regularisation.
Experiments performed on our \textit{concrete aggregate dataset} presented in this paper demonstrate the effectiveness of the proposed approach, outperforming the segmentation results achieved by purely supervised segmentation and standard consistency training.
}


\maketitle

%\sloppy % KAO: Sloppy spacing ensures non-overfull lines. Can be removed if this is not an issue.

%========================================================================================================
\section{Introduction}\label{Introduction}
Nowadays, concrete is the most dominant building material worldwide.
Concrete consists of a mixture of aggregate particles with a wide range of particle sizes (normally 0.1\,mm up to 32\,mm) and geometries (round, flat, ect.) which are dispersed in a cement paste matrix.
One important feature determining the quality and workability of fresh concrete is its stability which refers to the segregation behaviour of the concrete due to differences in specific weight or due to vibratory energy during the construction process \citep{Navarrete2016}.  
In this context, concrete whose aggregate distribution remains homogeneous over the height of the sample during the hardening phase is considered as stable while a sedimentation of the aggregate particles is an indicator for an unstable behaviour of the material. 
In order to assess the concrete stability a manual test method is used in which a hardened core of the target concrete is cut lengthwise and is visually examined by a human expert, evaluating the particle distribution. 
To overcome limitations resulting e.g.\ from errors in human judgement, the subjectivity of the evaluation, and from the fact that this process is labour-intensive, it it was suggested to develop automated systems to measure the concrete stability, e.g.\ based on image data of the sediment samples.  
However, so far only relatively simple approaches have been published, in which the aggregate is to be separated from the suspension based on manually defined intensity thresholds in order to derive information about the sedimentation behaviour \citep{Labi2007, Cotardo2017}.
In this paper, we propose a deep learning based approach for the segmentation of concrete aggregate in sedimentation images. 
%*******************************
\begin{figure}[ht]
\begin{center}
		\includegraphics[width=1.0\columnwidth]{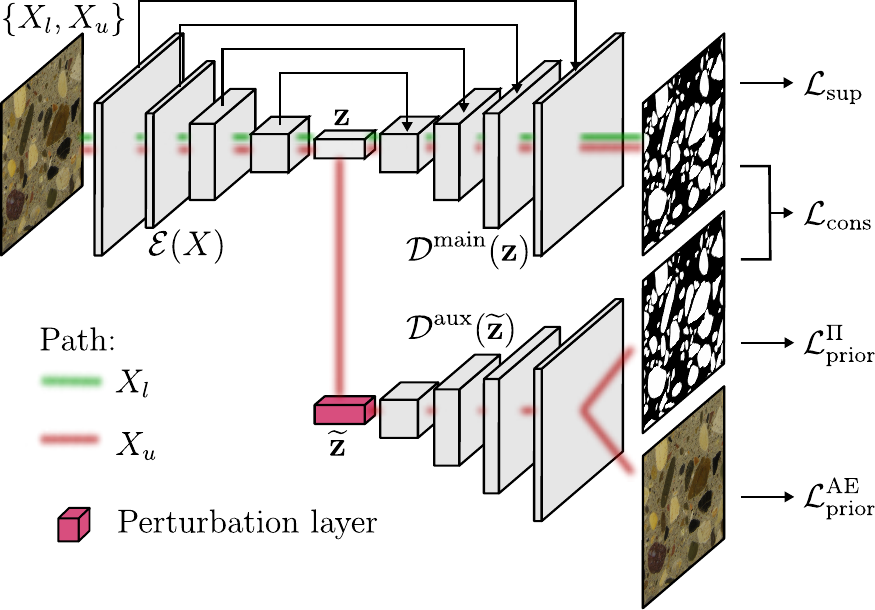}
	\caption{Overview of the proposed framework for semi-supervised semantic segmentation.
	}
\label{fig:overview}
\end{center}
\end{figure}
%*******************************
Typically, fully supervised approaches for image segmentation require large numbers of representative and annotated data in order to achieve high accuracies.
However, the generation of annotations, especially of pixel-wise reference labels, is highly tedious and time consuming. 
On the other hand, raw and unlabelled data can usually be acquired in abundance. 
The idea behind semi-supervised learning, therefore, is to leverage the large number of unlabelled data along with a limited amount of labelled data to improve the performance of deep neural networks. 
While several approaches for semi-supervised segmentation learning e.g.\ based on auto-encoder regularisation \citep{Myronenko2019}, entropy minimisation \citep{Kalluri2019}, consistency training \citep{Ouali2020}, or adversarial training \citep{Souly2017} have been proposed in the literature, the question of how to best incorporate unlabelled data is still an active problem in research. 

In this paper, we propose a novel framework for the semi-super\-vised training of deep learning networks for the segmentation of concrete particles. An overview of the framework can be seen in Fig.~\ref{fig:overview}.   
Building upon the concept of consensus regularisation \citep{Ouali2020}, we make the following contributions.\\
\textbf{1)} In a first step, we identify the weak spot of standard methods based on consistency training by presenting a theoretical derivation of their limitation which occurs when the data have imbalanced class distributions.\\
\textbf{2)} Having identified this limitation of the standard consensus regularisation as applied in existing work \citep{Ouali2020}, we propose a semi-supervised strategy using prior guidance to improve the segmentation performance (Sec.~\ref{sec:PriorGuidance}). 
In this context, we incorporate prior information into the training procedure in label space as well as in image space. 
More specifically, we make use of prior knowledge about the expected label distribution to supervise the label predictions of the unlabelled data and we introduce an image reconstruction loss based on an auto-encoder to learn the underlying distribution of the image data as additional regularisation of the encoder.\\
\textbf{3)} As an additional minor contribution we propose a light-weight architecture based on residual blocks and depthwise separable convolutions which achieves quality measures close to state-of-the-art while possessing significantly less parameters.\\
\textbf{4)} In order to train and to quantitatively evaluate the developed method we propose our \textit{concrete aggregate benchmark} consisting of high resolution images of cut concrete cores providing class labels on pixel-level. The dataset has been made freely available in the course of publication\footnote{ https://doi.org/10.25835/0027789}.

The remainder of this paper is structured as follows. We first provide a brief summary of related work in Sec.~\ref{sec:relatedWork}. A detailed identification of current limitations and a formal description of the proposed method is given in Sec.~\ref{sec:Methodology}.
In Sec.~\ref{sec:Setup} we present our new dataset and the evaluation of our method.
The paper is concluded in Sec.~\ref{sec:Conclusion}
%========================================================================================================
\section{Related work} \label{sec:relatedWork}
\subsection{Semantic Segmentation}
Semantic segmentation of images (called per- pixel classification in remote sensing) refers to the problem of assigning semantic labels to each pixel of an image.  
In this context, traditional approaches aim at finding a graph structure over image entities as e.g.\ pixels or superpixels by using a Markov Random Field (MRF) or Conditional Random Field (CRF) representation in order to capture context information.
Then, classifiers are employed to assign labels to the different entities based on carefully designed hand-crafted features \citep{LiSahbi2011, Sengupta2013, Coenen2017a}. \\
Nowadays, usually Convolutional Neural Networks (CNN) are applied for semantic segmentation in an end-to-end fashion. 
Pioneering work was presented by \citet{Long2015} who proposed a fully convolutional CNN for the per-pixel classification of images by replacing the fully connected layers of a standard CNN \citep{VGG19} by convolutional layers. 
In \citep{Noh2015}, transposed convolutions are proposed in order to create a learnable decoder which is added to the decoder, leading to an enhancement of the segmentation accuracy. 
Most of the current networks applied for semantic segmentation follow this encoder-decoder strategy. 
Skip-connections, also known as bypass connections \citep{Resnet2016} were firstly proposed by \citet{unet} for the task of semantic segmentation. The authors incorporated skip-connections between corresponding blocks of the encoder and the decoder in order to inject early-stage encoder feature maps to the decoder, which allows the subsequent convolutions to take place with awareness of the original feature maps, leading to better segmentation results at object borders.
In order to decrease the model size and the computational complexity of such encoder-decoder architectures, depthwise separable convolutions were proposed in \citep{MobileNet}, where the standard convolutional layers were replaced by operations which in a first step perform depthwise, i.e.\ per-channel convolutions in order to extract spatial features, followed by pointwise convolutions in order to learn cross-channel relations. 
In this work, we build upon the described state-of-the-art techniques for deep-learning based segmentation and propose a light-weight encoder-decoder architecture as basis for our framework for the semi-supervised segmentation of concrete aggregate.

\subsection{Semi-supervised segmentation}
In order to train semantic segmentation architectures, usually a large amount of pixel-wise annotated data representative for the classes to be extracted is required, which is tedious and expensive to obtain. 
Research on semi-supervised segmentation focusses on the question of how unlabelled data, which is typically easy to acquire in large amounts, can be used together with small amounts of labelled data to derive additional training signals in order to improve the segmentation performance. 

One line of research enriches the encoder-decoder structure of a supervised segmentation network by an additional auto-encoder which is trained in a self-supervised manner using the unlabelled data in order to improve the shared latent feature representation produced by the encoder \citep{Sedai2017,Myronenko2019}. 
The idea behind this strategy is to learn a common feature embedding for both tasks of semantic segmentation and reconstruction of the image. 
In this way, unlabelled data is used to add supplementary guidance and to impose additional constraints on the encoder part of the segmentation network. 
However, leveraging unlabelled data by providing guidance from auto-encoder reconstructions only considers the common distribution representing the image data but disregards reasoning on the level of semantic class labels of the unlabelled images.

As opposed to that, another strategy for making use of unlabelled data is based on entropy minimisation \citep{Kalluri2019, Wittich2020}, where additional training signals are obtained by maximising the network's pixel-wise confidence scores of the most probable class using unlabelled data. 
However, this approach introduces biases for unbalanced class distributions in which case the model tends to increase the probability of the most frequent and not necessarily of the correct classes.

In a semantic segmentation setting using adversarial networks, the segmentation network is extended by a discriminator network that is added on top of the segmentation and which is trained to discriminate between the class labels being generated by the segmentation network and those representing the ground truth labels.
By minimising the adversarial loss, the segmentation network is enforced to generate predictions that are closer to the ground truth and thus, they can be applied as additional training signal in order to improve the segmentation performance. 
In this context, the discrimination can be performed in an image-wise \citep{Luc2016} or pixel-wise \citep{Souly2017,Hung2018} manner. 
Since the adversarial loss can be computed without the need for reference labels once the discriminator is trained, the principles of adversarial segmentation learning are adapted for the semi-supervised setting to leverage the availability of unlabelled data \citep{Souly2017,Hung2018}.
However, learning the discriminator adds additional demands for labelled data and therefore might not reduce the need for such data in a way other strategies do.

Another line of research for semi-supervised segmentation is based on the consensus principle. 
In this context, \citet{Ouali2020} train multiple auxiliary decoders on unlabelled data by enforcing consistency between the class predictions of the main and the auxiliary decoders.
Similarly, in \citep{Peng2020} two segmentation networks are trained via supervision on two disjunct datasets and additionally, by applying a co-learning scheme in which consistent predictions of both networks on unlabelled data are enforced. 
Another approach based on consensus training is presented by \citet{Xiaomeng2018} and \citet{Zhang2020}, who use unlabelled data in order to train a segmentation network by encouraging consistent predictions for the same input under different geometric transformations. 
In this paper we argue that semi-supervised training based on the consensus principle leads to a problematic behaviour when dealing with imbalanced class distributions in the data. Tackling this problem, we propose a new strategy based on prior guidance in order to overcome this effect and to eventually improve the segmentation performance by making use of unlabelled data.

%========================================================================================================
\section{Methodology} \label{sec:Methodology}
\subsection{Problem statement}
CNN architectures for semantic image segmentation typically consist of an encoder $\mathcal{E}(X)$, which maps the input data $X$ to a latent feature embedding $\mathbf{z}$ by aggregating the spatial information across various resolutions, and of a decoder $\mathcal{D}(\mathcal{E}(X))=\mathcal{D}(\mathbf{z})$ which spatially upsamples the feature maps and finally applies a classifier to produce pixel-wise predictions $\hat{Y}$, usually at the same resolution as the input image.
In $\hat{Y}$, every pixel obtains a score $\hat{y}_i$ for each class $C_i \in \mathbf{C}$ with $i=1...N_C$, denoting the probability of the corresponding pixel to belong to the respective class.
In order to train such networks in a supervised manner, the reference label maps $Y$ are used to compute a pixel-wise loss $\mathcal{L}_\text{sup}(\hat{Y}, Y)$, which is backpropagated through the network via stochastic gradient descent (SGD) in order to optimise the network parameters. 
In this context, the availability of a sufficient amount of representative training data for which the reference labels are known is required for each class. 
In the absence of these labelled training data the neural network is likely to become overfitted, restricting the model's ability to generalise well and thus, restricting the performance of deep networks when applied to unseen data. 
Given a data set $X = \left\{X_l, X_u\right\}$, where $X_l$ are labelled examples possessing the reference labels $Y_l$ and $X_u$ are unlabelled examples for which no reference labels are available, the goal of this paper is to leverage the unlabelled data along with the labelled data for the training of a CNN in order to improve its performance compared to only using the labelled data. 
In this context, we regard the case where only a small number $N_l$ of labelled images but a large number $N_u$ of unlabelled data is available such that $N_u \gg N_l$. 
More specifically, we train a fully convolutional encoder-decoder CNN for the task of concrete aggregate segmentation. 
However, we point out that the proposed framework can be applied to any encoder-decoder based network. 

\subsection{Semi-supervision using consensus regularisation}
In this work, we build upon an encoder-decoder network as described above.
In the remainder of this paper, we refer to the decoder performing the classification as the \textit{main} decoder $\mathcal{D}^\text{main}(\mathbf{z})$ and to the predicted label maps as $\hat{Y}^\text{main}$.
In addition, we introduce an \textit{auxiliary} decoder $\mathcal{D}^\text{aux}(\widetilde{\mathbf{z}})=\hat{Y}^\text{aux}$. 
Both decoders make use of the shared encoder $\mathcal{E}(X) = \mathbf{z}$ to predict the target label maps. 
While the main decoder is trained in a supervised manner on the labelled data $X_l$ using the corresponding label maps $Y_l$ to compute the loss $\mathcal{L}_\text{sup}(\hat{Y}_l^\text{main}, Y_l)$, the auxiliary decoder is trained on the unlabelled data $X_u$ by enforcing consistency between predictions of the \textit{main} decoder and the \textit{auxiliary} decoder.
In this context, the training objective is to minimise the consensus loss $\mathcal{L}_\text{cons}(\hat{Y}_u^\text{main}, \hat{Y}_u^\text{aux})$, which gives a measure of the discrepancy between the predictions of the \textit{main} and the \textit{auxiliary} decoder.
In order to ensure diversity between both decoders, a perturbed version $\widetilde{\mathbf{z}}$ of the latent representation $\mathbf{z}$ with $\widetilde{\mathbf{z}} = \mathcal{F}(\mathbf{z})$, using a perturbation function $\mathcal{F}(\cdot)$, is fed to the \textit{auxiliary} decoder while the uncorrupted representation $\mathbf{z}$ is used as input for the \textit{main} decoder.
This procedure of consensus regularisation for semi-supervised segmentation is founded on the rationale that the shared encoder's representation can be enhanced by using the additional training signal obtained from the unlabelled data, acting as additional regularisation on the encoder \citep{Ouali2020, Peng2020}. 
Based on the consensus principle \citep{Chao2016}, enforcing an agreement between the predictions of multiple decoder branches restricts the parameter search space to cross-consistent solutions and thus, improves the generalisation of the different models.  
Furthermore, the perturbations aim at enforcing invariance to small deviations in the latent representation of the data.

\subsection{The \textit{blind spot} of the consensus principle} \label{sec:BlindSpot}
In this section, we present a theoretically founded derivation of the limitations behind semi-supervised training using the consensus principle. 
In an unsupervised training setup based on the consensus principle as described above and as applied in the literature \citep{Ouali2020, Peng2020}, the training signal is computed based on the discrepancy between the predictions of two or more distinct models. 
Consequently, knowledge about the reference labels is not required in order to compute the consensus training loss $\mathcal{L}_\text{cons}$, which is the reason why also unlabelled data can be leveraged for training. 
Instead, a training signal is produced if the models disagree on the prediction and no training signal is produced if the models agree on the prediction, regardless of the fact whether the prediction is correct or not. 
In this context, the pixel-wise class predictions of each model can be categorised by an unknown binary state variable $\mathbf{s} \in \left\{s^+, s^-\right\}$ signalising if the pixel is classified correctly ($s^+$) or incorrectly ($s^-$).
The \textit{blind spot} of consensus training occurs in cases where the models agree on their prediction, so that consequently no training signal is produced, even though the predictions are incorrect ($\mathbf{s} = s^-$), i.e. they do not match the actual class label.  
In this paper, we argue that the effect of the \textit{blind spot} just described leads to an unfavourable guidance by the consensus principle, provided a data set possesses an imbalanced label distribution, i.e.\ it consists of data in which one or more classes occur more frequently compared to others.\\
The joint probability of a pixel to belong to the reference class $C_i$ and to be classified either correctly or incorrectly can be expressed as
\begin{equation} 
P(\mathbf{s}, C_i) = P(\mathbf{s}|C_i) \cdot P(C_i). \label{eq:JointProb}
\end{equation}
In this expression, $P(C_i)$ is the prior probability of the pixel to belong to the reference class $C_i$ and can be represented by the proportion of the respective class in the data. 
Assuming that the probability, whether a classifier is able to determine the correct class for a pixel or not, is independent of the actual class of the pixel, leads to the state $\mathbf{s}$ and the class $\mathbf{C}$ to be independent variables and, therefore, simplifies the conditional probability $P(\mathbf{s}|\mathbf{C})$ to read
\begin{equation} 
P(\mathbf{s}|C_i) = P(\mathbf{s}) \quad \forall i. \label{eq:CondProb}
\end{equation}
To gain further insights into the probabilistic behaviour of predictions leading to the \textit{blind spot} of the consensus principle, the case where $\mathbf{s}:=s^-$ is investigated further.
In order to introduce the predicted class $\hat{C}_k$ into the probabilistic formulation, the joint probability of the reference and the predicted class, and the state $\mathbf{s}=s^-$ is formulated as
\begin{equation}
\begin{aligned}
P(\hat{C}_k, C_i, s^-) 	& = P(\hat{C}_k|s^-, C_i) \cdot P(s^-, C_i) \quad \text{with } k\neq i \\
												& = P(\hat{C}_k|s^-, C_i) \cdot P(s^-) \cdot P(C_i)
\end{aligned} \label{eq:JointProbPixel}
\end{equation}
For simplification, we assume that the conditional probability $P(\hat{C}_k|s^-, C_i)$ of the predicted class $\hat{C}_k$ is independent of the actual class $C_i$ (although in practice, this assumption does not always hold true, for example an instance of the class \textit{dog} might be more likely misclassified as \textit{cat} than e.g.\ as \textit{bird} etc).
With an overall number of classes $N_C$, this simplification leads to 
\begin{equation}
P(\hat{C}_k|s^-, C_i) = P(\hat{C}_k|s^-) = \frac{1}{N_C - 1} \quad \text{with } k \neq i, 
\end{equation}
and therefore Eq.~\ref{eq:JointProbPixel} simplifies to
\begin{equation}
P(\hat{C}_k, C_i, s^-) = \frac{1}{N_C - 1} \cdot P(s^-) \cdot P(C_i). \label{eq:JointPredRefState}
\end{equation}

The probability, that two classifiers $\mathcal{D}^\text{main}$ and $\mathcal{D}^\text{aux}$ agree on the same but incorrect class label $\hat{C}_k$ such that  $\mathbf{s}^\text{main}=\mathbf{s}^\text{aux}=s^-$ and $\hat{C}_k^\text{main}=\hat{C}_k^\text{aux}\neq C_i$ occurs at a pixel with the actual class $C_i$, can be expressed by the joint probability 
\begin{equation}
\begin{aligned}
& P(\hat{C}_k^\text{main}, \mathbf{s}^\text{main}=s^-, \hat{C}_k^\text{aux}, \mathbf{s}^\text{aux}=s^-, C_i) \\
= & P(\hat{C}_k^\text{main}, \mathbf{s}^\text{main}=s^-|\hat{C}_k^\text{aux}, \mathbf{s}^\text{aux}=s^-, C_i) \\
& \cdot P(\hat{C}_k^\text{aux}, \mathbf{s}^\text{aux}=s^-, C_i)
\end{aligned} \label{eq:ProbBlindSpotLong}
\end{equation}
Considering the two classifiers $\mathcal{D}^\text{main}$ and $\mathcal{D}^\text{aux}$ as independent from each other allows to simplify the conditional probability in Eq.~\ref{eq:ProbBlindSpotLong} according to
\begin{equation} 
\begin{aligned}
& P(\hat{C}_k^\text{main}, \mathbf{s}^\text{main}=s^-|\hat{C}_k^\text{aux}, \mathbf{s}^\text{aux}=s^-, C_i)\\
= & P(\hat{C}_k^\text{main}, \mathbf{s}^\text{main}=s^-) \\
= & P(\hat{C}_k^\text{main}| \mathbf{s}^\text{main}=s^-) \cdot P(\mathbf{s}^\text{main}=s^-)\\
= & \frac{1}{N_C-1} \cdot P(s^-)
\end{aligned} \label{eq:CondBlind}
\end{equation}
By substituting Eq.~\ref{eq:ProbBlindSpotLong} with Eqs.~\ref{eq:JointPredRefState} and \ref{eq:CondBlind}, the probability of a \textit{blind spot} to occur results in
\begin{equation}
\begin{aligned}
& P(\hat{C}^\text{main}=\hat{C}^\text{aux}, \mathbf{s}^\text{main}=\mathbf{s}^\text{aux}=s^-, C_i) \\
= & \sum_{\forall k\neq i} \frac{P(s^-)^2}{(N_C-1)^2} \cdot P(C_i) =\frac{P(s^-)^2}{N_C-1} \cdot P(C_i)\\
%= & \frac{P(s^-)^2}{N_C-1} \cdot P(C_i).
\end{aligned} \label{eq:BlindSpot}
\end{equation}
Finally, according to Eq.\ref{eq:BlindSpot}, the probability of the occurrence of a \textit{blind spot} during consensus regularisation solely varies in dependency of the prior probability of the reference class $C_i$. 
In case of data exhibiting an imbalanced label distribution such that $\exists i(P(C_i) > P(C_k) \land i \neq k)$, i.e.\ if there exist one or more classes which appear more often than other classes, the probability of a \textit{blind spot} to occur for instances of that class is larger compared to other classes and therefore, statistically fewer training signals are produced for incorrect predictions of the respective majority classes.  
As a consequence, consensus regularisation systematically favours the prediction of more common classes by introducing a bias within the consensus loss $\mathcal{L}_\text{cons}$ to the training procedure.

\subsection{Consensus regularisation with prior guidance} \label{sec:PriorGuidance}
In this paper, we propose a strategy to overcome the unintended effect of consensus regularisation described in Sec.~\ref{sec:BlindSpot} by making use of prior information which is exploited for further guidance of the semi-supervised training procedure.
In this context, on the one hand, we compute the class distribution $\Pi(Y_l)$ within the labelled training data $Y_l$ in order to introduce an additional loss $\mathcal{L}_\text{prior}^\Pi(\hat{Y}_u, \Pi(Y_l))$ to the training of the proposed CNN which enforces the network to produce a label distribution of the predicted label maps that corresponds to the class distribution of the training data.
By doing so, we aim to counteract the biasing effect introduced by the consensus principle in $\mathcal{L}_\text{cons}$ negatively affecting the prediction of less common classes.
On the other hand, the image data itself can be considered and leveraged as prior information.
To this end, we add an additional output $\hat{X}$ to the \textit{auxiliary} decoder $\mathcal{D}^\text{aux}$ which aims at reconstructing the input image $X$ itself in order to introduce additional prior guidance using auto-encoder regularisation. 
By doing so, we build upon the idea proposed in \citep{Sedai2017,Myronenko2019} and introduce an auto-encoder to the segmentation network in order to regularise the shared decoder and to impose additional constraints on its parameters.
To this end, we add a reconstruction loss $\mathcal{L}_\text{prior}^\text{AE}(\hat{X}_u^\text{aux}, X_u)$ which measures the discrepancy between the input image an the image reconstructed by the auto-encoder.
In this way, we aim at leveraging the inherent feature similarity of the large number of unlabelled images by enforcing the encoder to learn a latent feature representation of the auto-encoding model. 
An overview on the complete framework proposed in this paper for the task of semi-supervised segmentation is shown in Fig.~\ref{fig:overview}.
As depicted, both decoders share the same encoder. During training, both, labelled and unlabelled data $X_l$ and $X_u$ is passed through the \textit{main} decoder while only the unlabelled data $X_u$ is processed by the \textit{auxiliary} decoder.
The training objective is to minimise the overall training loss 
\begin{equation}
\begin{aligned}
\mathcal{L} = &\ \mathcal{L}_\text{sup}(\hat{Y}_l^\text{main}, Y_l) + \omega_1 \mathcal{L}_\text{cons}(\hat{Y}_u^\text{main}, \hat{Y}_u^\text{aux})\\
& + \omega_2 \mathcal{L}_\text{prior}^\Pi(\hat{Y}_u^\text{aux}, \Pi(Y_l)) + \omega_3 \mathcal{L}_\text{prior}^\text{AE}(\hat{X}_u^\text{aux}, X_u). \label{eq:totalLoss}
\end{aligned}
\end{equation}
A detailed description of the individual components of the loss formulation is given in the subsequent paragraphs. 

\subsubsection*{Supervised loss:}
For a labelled training sample $X_l$ and $Y_l$, the segmentation network $\mathcal{D}^\text{main}(\mathcal{E}(X_l))$ is trained using the supervised loss $\mathcal{L}_\text{sup}(\hat{Y}_l^\text{main}, Y_l)$. For $\mathcal{L}_\text{sup}$, the weighted mean squared error (MSE) loss as proposed in \citep{Coenen2019} is computed from the predicted label maps $\hat{Y}_l^\text{main}$ and the reference label maps ${Y}_l$.

\subsubsection*{Consensus loss:}
The consensus loss $\mathcal{L}_\text{cons}(\hat{Y}_u^\text{main}, \hat{Y}_u^\text{aux})$ is an unsupervised loss and measures the discrepancy between the \textit{main} decoder's predictions $\hat{Y}_u^\text{main}$ and those of the \textit{auxiliary} decoder $\hat{Y}_u^\text{aux}$ for the unlabelled training exampled $X_u$. As distance measure, the MSE is used in this work. 

\subsubsection*{Prior loss:}
The prior loss $\mathcal{L}_\text{prior}^\Pi(\hat{Y}_u^\text{aux}, \Pi(Y_l))$ is based on the difference between the class distribution of the predicted label maps $\Pi(\hat{Y}_u^\text{aux})$ and the prior class distribution $\Pi(Y_l)$ derived from the labelled training data. 
In order to compute $\Pi(Y_l)$, we calculate the proportion of pixels of each class $C_i$ with $i=1...N_C$ w.r.t. to the overall number of pixels for each image in $Y_l$. 
We represent $\Pi(Y_l)$ by the average class proportions $\mu_i$ and the standard deviation $\sigma_i$ across the whole training set $X_l$. 
Given the class distribution $\Pi(Y_l)$ determined a priori, the prior loss is computed according to
\begin{equation}
\mathcal{L}_\text{prior}^\Pi(\hat{Y}_u^\text{aux}, \Pi(Y_l)) = \frac{1}{N_C} \sum_{i=1}^{N_C} \left(\frac{p_i(\hat{Y}_l^\text{aux}) - \mu_i}{2\sigma_i} \right)^2. \label{eq:PriorLoss}
\end{equation}
In Eq.~\ref{eq:PriorLoss}, $p_i(\hat{Y}_l^\text{aux})$ denotes the proportion of pixels belonging to class $C_i$ of the predicted label map $\hat{Y}_l^\text{aux}$.
This loss enforces the \textit{auxiliary} decoder to predict label maps inheriting the label distribution from the training data and therefore acts as counterweight to the bias towards predicting more frequent classes introduced by the consensus loss.

\subsubsection*{Auto-encoder loss:}
The loss $\mathcal{L}_\text{prior}^\text{AE}(\hat{X}_u, X_u)$ is a self-super\-vised loss and is computed for the unlabelled images based on the discrepancy of the auto-encoder output $\hat{X}_u$ of the \textit{auxiliary} decoder and the input image $X_u$. 
In this work, the MSE is computed as distance measure to compute the auto-encoder loss.
Introducing this loss allows for additional training guidance using the principles of auto-encoder regularisation.

The parameters $\omega_1$, $\omega_2$ and $\omega_3$ in Eq.~\ref{eq:totalLoss} act as factors to weigh the individual components of the overall loss of Eq.~\ref{eq:totalLoss} w.r.t. each other. 
It has to be noted, that only the labelled examples are used to train the \textit{main} decoder as only the supervised loss is backpropagated through $\mathcal{D}^\text{main}$, while the unlabelled data is leveraged for the training of the \textit{auxiliary} decoder $\mathcal{D}^\text{aux}$ in an un-/self-supervised manner, respectively.

%========================================================================================================
\section{Experiments} \label{sec:Setup}

\subsection{Test data}
\begin{figure}[H]
\begin{center}
		\includegraphics[width=1.0\columnwidth]{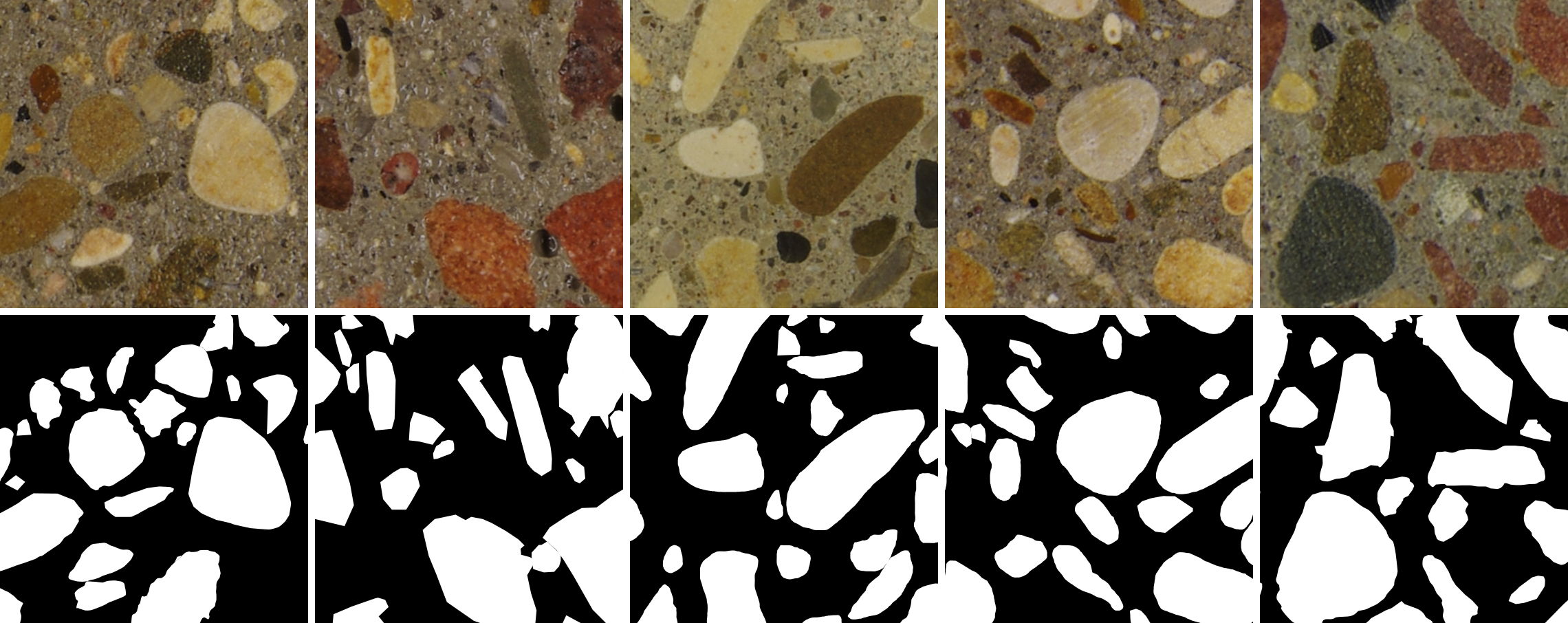}
	\caption{Examples of our proposed data set. The top row shows exemplary images of concrete and the bottom row shows the reference labels with \textit{white} and \textit{black} representing the classes \textit{aggregate} and \textit{suspension}, respectively.}
\label{fig:Dataset}
\end{center}
\end{figure}

\begin{figure}[H] 
\centering
%\hspace{-1.0cm}
\subfloat[Histogram w.r.t. the max. particle diameter (mm)] {\includegraphics[width=0.72\columnwidth]{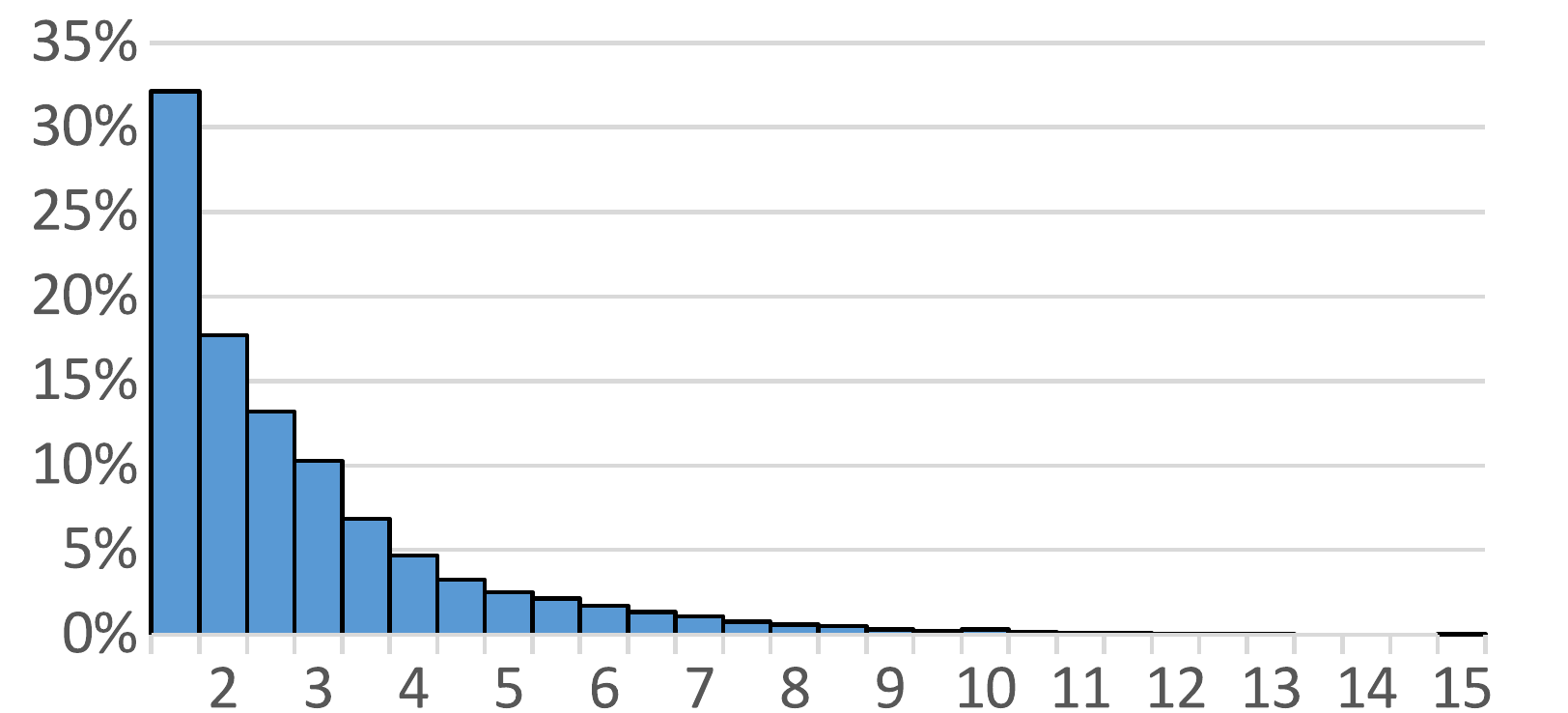}\label{fig:ParticleDiameter}} 
\hspace{0.0001cm}
\subfloat[Cumulative histogram w.r.t. particle area ($\text{mm}^2$)] {\includegraphics[width=0.72\columnwidth]{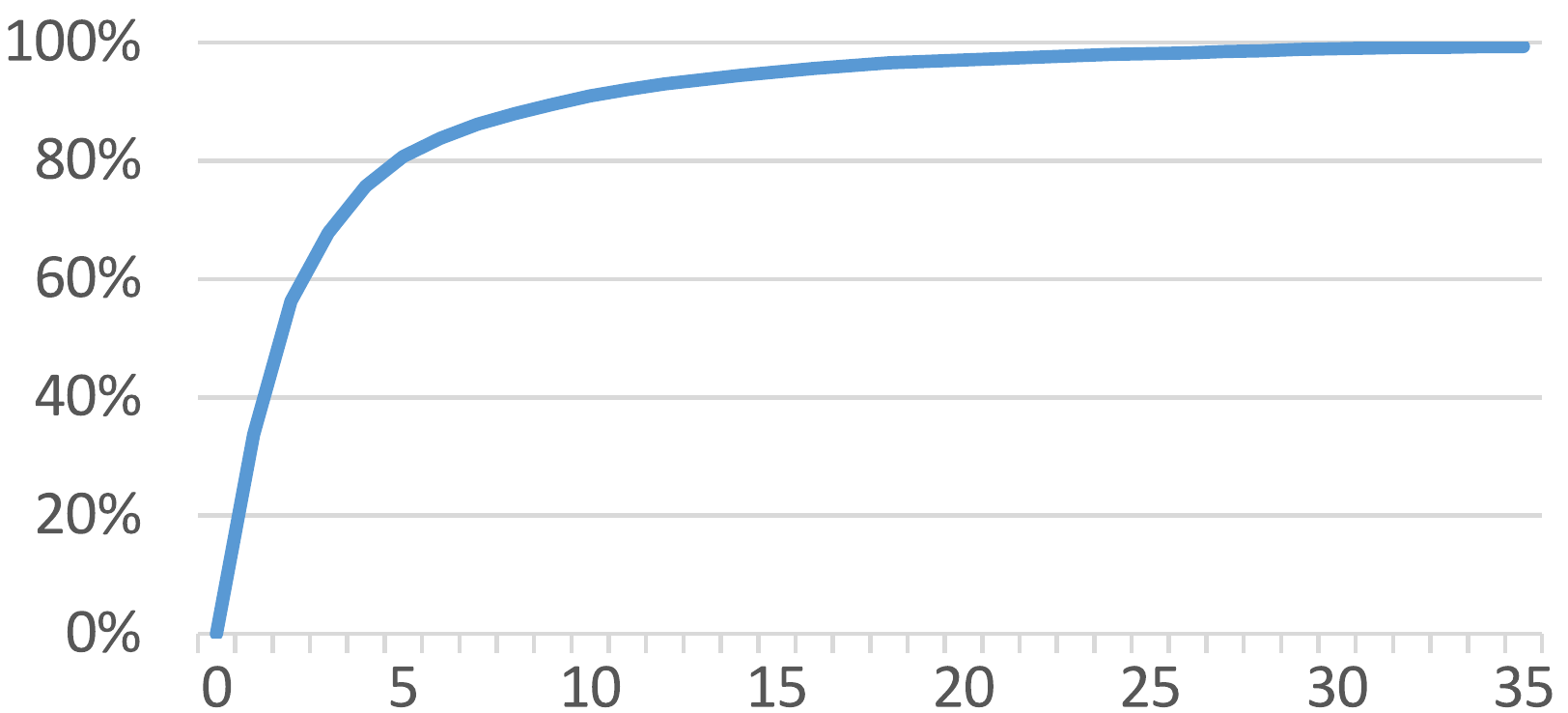}\label{fig:ParticleArea}} 

\caption{Distribution of the particles in the proposed data set w.r.t. their maximum diameter and their area.}
\label{fig:ParticleStatistics} 
\end{figure}

To evaluate the proposed approach for semi-supervised segmentation and its applicability for the segmentation of concrete aggregate we provide a new data set in the course of this paper.
To this end, high resolution images were acquired from 40 different concrete cylinders, cut lengthwise as to display the particle distribution in the concrete, with a ground sampling distance of 30\,$\mu m$.
Each sedimentation image is subdivided into 36 tiles of size 448x448$\text{px}^2$. 
At the time of submission, 612 tiles belonging to images from 17 different sedimentation pipes have been  annotated by manually associating one of the classes \textit{aggregate} or \textit{suspension} to each pixel. 
The remaining images are used as unlabelled data for the semi-supervised segmentation training proposed in this paper. 
With 36.2\% of all annotated pixels belonging to the class \textit{aggregate} and 63.8\% of the data being associated to the class \textit{suspension}, the data contains an imbalanced class distribution with the class \textit{aggregate} representing the minority class. 
As a consequence, our data set presents a suitable test environment for our proposed semi-supervised segmentation framework tackling the problems of consensus-learning that occur in the context of imbalanced label distributions in the data.
An overview of the statistics of the dataset is given in Tab.~\ref{tab:Dataset}. 
Fig.~\ref{fig:Dataset} shows five exemplary tiles and their annotated label masks. 
The diversity of the appearance of both, \textit{aggregate} and \textit{suspension} can be noted. 
%=====================================================
\begin{table}[H]
	\centering
	%\scriptsize
	\caption{Statistics of our proposed concrete aggregate data set.}
		\begin{tabular}{|l| c c c|} \hline
		& labelled & unlabelled & total\\ \hline
		No. of images & 17 & 23 & 40\\
		No. of tiles & 612 & 828 & 1440 \\ \hline
		\end{tabular}	
\label{tab:Dataset}
\end{table}  
%=====================================================
%*******************************
%******************************* 
In Fig. \ref{fig:ParticleStatistics}, the distribution of the particles in dependency on their sizes is depicted. 
The variation of the size of the particles contained in the data set ranges up to 15\,mm of maximum particle diameter. However, the majority of particles, namely more than 50\% exhibit a maximum diameter of less then 3\,mm (100px). 
As a consequence, approximately 80\% of the particles possess an area of 5\,$\text{mm}^2$ or less. 
%==================================================================================
%
%
%==================================================================================

It has to be noted that particles with a size less then 20px are barely distinguishable from the suspension and are therefore not contained in the reference data. 

\subsection{Architectures}
In order to evaluate the effect of the proposed framework for semi-supervised segmentation we make use of two different fully convolutional segmentation architectures.
However, we point out that the proposed strategy for semi-supervised segmentation learning can be adapted to any arbitrary encoder-decoder network structure since its applicability is not restricted to any specific architecture. 
The first architecture that is used in the experiments is the \textit{Unet} proposed by \citet{unet}, which is an encoder-decoder architecture with approx.\ 31 Mio.\ learnable parameters, which, thus, represents a rather heavy-weight network structure.

\begin{figure}[H]
\begin{center}
		\includegraphics[width=1.0\columnwidth]{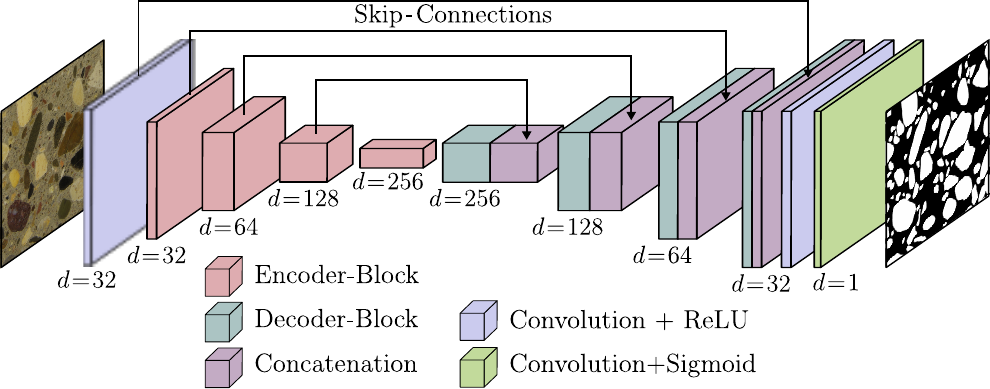}
	\caption{High-level overview on our proposed R-S-Net architecture. The depth of the feature maps is denoted by $d$.}
\label{fig:Architecture}
\end{center}
\end{figure}

\begin{figure}[H]
\begin{center}
		\includegraphics[width=1.0\columnwidth]{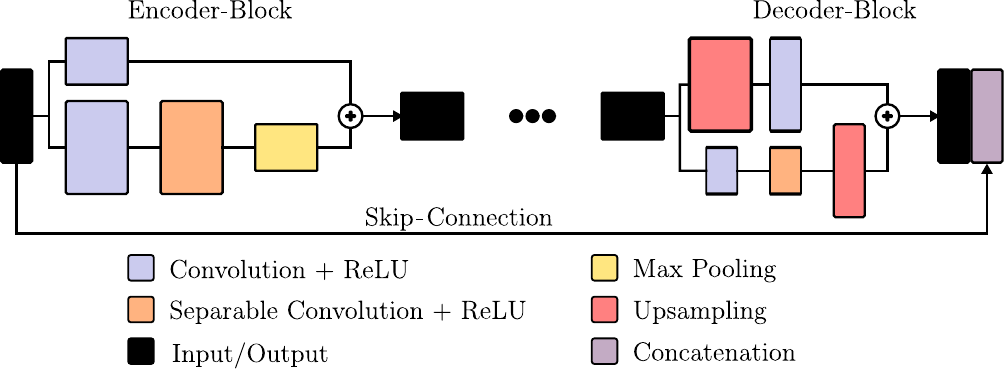}
	\caption{Structure of the R-S-Net encoder and decoder blocks. 
	%applied in our proposed R-S-Net architecture.
	}
\label{fig:Blocks}
\end{center}
\end{figure}

In addition, we propose the \textit{R-S-Net} (\textbf{R}esidual depthwise \textbf{S}epar\-able convolutional \textbf{Net}work), a lightweight CNN with approx.\ 1.9 Mio.\ parameters, thus more than 16 times fewer parameters compared to the Unet.
A high-level overview of the used encoder-decoder network architecture is shown in Fig.~\ref{fig:Architecture}. 
Note that for reasons of simplicity, Fig.~\ref{fig:Architecture} only depicts the architecture of the encoder and the \textit{main} decoder.
The \textit{auxiliary} decoder used for the semi-supervised training is identical to the \textit{main} decoder of the respective architecture, except that no skip-connections are used and the latent feature map produced by the encoder undergoes stochastic permutations (described later) before it is fed to the \textit{auxiliary} decoder.
The additional decoder branch leads to an overhead of parameters during training, however, the \textit{auxiliary} decoder is only used during training; for inference, only the \textit{main} decoder is used.
%(although this is arguably not a very strong argument, as the main complexity is that of training).
% na ja, aber zeitintensiv ist ja gerade das Training - das Argument ist also nciht eben riesig wichtig, oder? Daher der Nachsatz in Klammern ... 
%*******************************
%*******************************
The input to the CNN is a three-channel colour image of a concrete sample profile.
The encoder $\mathcal{E}$ consists of a convolutional layer, followed by four encoder-blocks.
The decoder is symmetric to the encoder and consists of four decoder-blocks followed by convolutional layers. 
% bitte prüfen, ob der Plural ("layers") hier richtig ist - ich habe das "s" hinzugefügt. Wenn das nciht stimmt, sollte es "a convolutional layer" heißen.
Both convolutional layers use filters with a kernel size of 3x3 and ReLU as non-linear activation function. 
Skip-connections are are used between corresponding encoder-decoder-blocks by concatenating the outputs of the encoder-blocks to the outputs of the decoder-blocks of the same spatial size.
The final output, i.e.\ the segmentation map is produced by an additional convolutional layer using a 1x1 filter kernel and a sigmoid activation function. 
Details on the structure of the encoder- and decoder-blocks are shown in Fig.~\ref{fig:Blocks} and are explained in the following paragraphs.
%*******************************
%*******************************
\subsubsection*{Encoder-block}
Each encoder-block consists of a residual convolution module, which takes a feature map of size $m \times n$ as input and which returns a feature map with depth $d$ and with spatial size of $m/2 \times n/2$ as output.
Inside each encoder block, two intermediate representations are computed from the initial feature map.
The first representation is produced by a convolutional layer using a kernel size of 1 and a stride of 2, and the second one is computed by a sequence of a convolutional layer followed by a depthwise separable convolution layer \citep{MobileNet}, both using kernel size 3x3 and stride 1, and downsampled using max. pooling with kernel size 2x2 and stride 2. As non-linear activation function, ReLU, is applied in each of the convolutional layers.
As output of each block, the element-wise sum of both intermediate representations is returned. 

\subsubsection*{Decoder-block}
\sloppy
Similar to the encoder-block, the decoder\-block processes the input in a two-stream path and returns the element-wise sum of the output of both streams. 
In the first stream, the input is upsampled by a factor of 2, followed by a convolutional layer using filters with kernel size 1x1. 
The second stream consists of a sequence of a convolutional layer followed by a depthwise separable convolution (both using kernel sizes of 3x3) and an upsampling layer.

\subsubsection*{Perturbation layer}
Similar to \citet{Ouali2020}, we apply perturbations $\mathcal{F}$ to the latent variable $\mathbf{z}$ produced by the encoder to obtain the perturbed feature map $\tilde{\mathbf{z}} = \mathcal{F}(\mathbf{z})$, which is then fed to the \textit{auxiliary} decoder.
The perturbation layer applies two feature based perturbations leading to $\mathcal{F}(\mathbf{z}) = \mathcal{F}_\text{Drop}(\mathcal{F}_\text{Noise}(\mathbf{z}))$.
In $\mathcal{F}_\text{Noise}$, a noise tensor $\mathbf{N}$ is uniformly sampled in the range of $(-0.3,0.3)$ and is injected to the encoder's output:
\begin{equation}
\mathcal{F}_\text{Noise}(\mathbf{z}) = \left(\mathbf{z} \odot \mathbf{N}\right) + \mathbf{z}, 
\end{equation}
Here, $\odot$ denotes an element-wise multiplication of two tensors.
In $\mathcal{F}_\text{Drop}$, a proportion of the feature map with the highest activations is set to zero. To this end, a threshold $\gamma$ is randomly drawn from the uniform distribution in the range of $(0.6,0.9)$. 
After channel-wise normalising of the feature map $\mathbf{z}$ resulting in $\mathbf{z}'$, each entry of $\tilde{\mathbf{z}}$ is set to 0 whose value in $\mathbf{z}'$ exceeds the threshold $\gamma$.

\subsection{Evaluation strategy and training}
In order to assess the impact of the proposed method for semi-supervised segmentation, different variants for the network are defined, each considering different components and loss functions of the framework presented in this paper.

\subsubsection*{Base:} In the baseline settings $\text{Unet}_\text{base}$ and $\text{R-S-Net}_\text{base}$, the performance of the two baseline architectures is evaluated, i.e.\ in this set the \textit{auxiliary} decoder is not used during training and consequently, training is done in a standard supervised manner.

\subsubsection*{Consensus:} In this setting, denoted as $\text{Unet}_\text{cons}$ and $\text{R-S-Net}_\text{cons}$, semi-supervised training is done by considering the \textit{auxiliary} branch and applying the consensus loss $\mathcal{L}_\text{cons}$. In this way, the effect of considering unlabelled data following the consensus principle as proposed in \citep{Ouali2020} can be assessed.

\subsubsection*{Consensus+prior:} The settings $\text{Unet}_\text{full}$ and $\text{R-S-Net}_\text{full}$ make use of the complete framework presented in this paper by considering the full formulation of Eq.~\ref{eq:totalLoss}. 
In this work, the weights $\omega_{1-3}$ are set to 1. A properly defined individual weighting of the different losses might further improve the performance of the network, but is not evaluated in the scope of this paper. 

\subsubsection*{Training:} The CNNs used in the different variants of the proposed framework are trained from scratch.
The convolutional layers are initialised using the \textit{He} initialiser \citep{He2015}.
The networks are trained using the Adam optimizer \citep{Adam2015}, a variant of stochastic mini-batch gradient descent with momentum, using the exponential decay rate for the $1^\text{st}$ moment estimates $\beta_1 = 0.9$ and for the $2^\text{nd}$ moment estimates $\beta_2 = 0.999$.  
We apply weight regularisation on the convolutional layers using L2 penalty with a regularisation factor of $10^{-5}$.
A mini-batch size of 4 is applied, meaning that each mini-batch consists of four labelled and four unlabelled training images. 
We use an initial learning rate of $ 10^{-3}$ and decrease the rate by a factor of $10^{-1}$ after 25 epochs with no improvement in the training loss and 
train each setting for 500 epochs.
In order to get insights into the effect of the amount of annotated training data on the quality of the segmentation results, we vary the number of training images in the conducted experiments. 
We define a minimum setting $T_1$, in which only one tile of each of the 17 annotated sedimentation pipes is used for the supervised training part of the segmentation framework. In $T_3$, $T_5$, and $T_{10}$, three, five, and ten annotated tiles of each sedimentation pipe, respectively, are used for training.
The values for $\mu_i$ and $\sigma_i$ of Eq.\ref{eq:PriorLoss} are computed from the individual training sets.
In all variants, we make use of all 828 non-annotated images to compute the losses of the non-supervised part of the framework.
% heißt das, dass Du keinen validation set nutzt? Ist das nciht eher "gefährlich"?

\subsubsection*{Evaluation metrics:}
The evaluation of our proposed method is based on all annotated concrete aggregate tiles that have not been used for training.
We determine values for the overall accuracy (OA) of the segmentation, as well as class-wise values for \textit{recall}, \textit{precision}, and \textit{$\text{F}_1$-score} according to:
\begin{equation}
\text{Recall}\ [\%] = 100 \cdot \frac{TP}{TP+FN}
\end{equation}
\begin{equation}
\text{Precision}\ [\%] = 100 \cdot \frac{TP}{TP+FP}
\end{equation}
\begin{equation}
\text{F}_1\text{-score}\ [\%] = 2 \cdot \frac{\text{Precision} \cdot \text{Recall}}{\text{Precision} + \text{Recall}}
\end{equation}
In these equations, TP (\textit{true positives}) denotes the number of correctly classified pixels per class, FN (\textit{false negative}) is the number of pixels of that class that are erroneously classified  and FP is the number of pixels that are erroneously classified as the class under consideration (\textit{false positives}).
The $F_1$-score is the harmonic mean of precision and recall and, thus, is not biased towards more frequent classes.
In addition to the OA, which can be biased for imbalanced class distributions, we report the \textit{Mean $\text{F}_1$-score} ($\text{MF}_1$) of both classes.

%========================================================================================================
\subsection{Results} \label{sec:Evaluation}
In Tab.~\ref{tab:Results} the OA, the $MF_1$-score, and the class-wise values for \textit{precision}, \textit{recall} and \textit{$F_1$-score} achieved by the different variants of the Unet and the R-S-Net based on the $T_1$ setup are shown.
For a visual comparison, Fig.~\ref{fig:QualRes} shows examplary qualitative results achieved by the R-S-Net using the various settings of the framework.
%=====================================================
\begin{table*}[ht]
	\centering
	%\scriptsize
	\caption{Quantitative results for the different settings of the proposed framework achieved for the $T_1$ training setup.}
		\begin{tabular}{l| c c | c c c | c c c} \hline
		& & & \multicolumn{3}{c|}{Aggregate} & \multicolumn{3}{c}{Suspension} \\
		values in [\%]& OA & $MF_1$ & Recall & Precision & $F_1$ & Recall & Precision & $F_1$ \\ \hline \hline
		$\text{Unet}_\text{base}$ \citep{unet} 		& 88.0 & 86.9	& \textbf{85.0}	& 81.4	& 83.2	& 89.6	& \textbf{91.7}	& 90.6 \\
		$\text{Unet}_\text{cons}$ 								& 88.5 & 86.6	& 72.6	& \textbf{93.1}	& 81.5	& \textbf{97.1}	& 86.8	& 91.7 \\
		$\text{Unet}_\text{full}$ 								& \textbf{90.5} & \textbf{89.4}	& 82.4	& 89.6	& \textbf{85.9}	& 94.8	& 90.9	& \textbf{92.8} \\ \hline
		$\text{R-S-Net}_\text{base}$ 							& 85.8 & 84.7	& \textbf{83.6}	& 77.6	& 80.5	& 87.0	& \textbf{90.8}	& 88.9 \\
		$\text{R-S-Net}_\text{cons}$ 							& 89.4 & 87.9	& 76.9	& \textbf{91.5}	& 83.5	& \textbf{96.1}	& 88.5	& 92.2 \\
		$\text{R-S-Net}_\text{full}$ 							& \textbf{89.8} & \textbf{88.5}	& 80.1	& 89.7	& \textbf{84.6}	& 95.0	& 89.9	& \textbf{92.4} \\ \hline
		\end{tabular}	
\label{tab:Results}
\end{table*}  
%=====================================================
\subsubsection*{Base:} 
The OA that is achieved by training the two considered architectures in a purely supervised manner, i.e.\ without the consideration of additional unlabelled data during training, results in 85.8\% for the lightweight R-S-Net and in 88.0\% for the Unet.
Similarly, the $MF_1$-score of the \textit{base} architectures (86.9\%) is larger for the Unet compared to the result achieved by the R-S-Net (84.7\%). 
Consequently, applying purely supervised training to learn the mapping from the image to the label space leads to a better performance of the Unet over the light-weight R-S-Net.
As can be seen from Fig.~\ref{fig:QualRes} for the \textit{base} setting, a relatively large proportion of the FN \textit{aggregate} classifications, i.e.\ \textit{aggregate} pixels that were erroneously classified as \textit{suspension}, belong to boundaries of the individual aggregate particles.
In comparison, the FP \textit{aggregate} classifications, i.e.\ pixel that were erroneously associated to aggregate particles mostly appear as larger connected segments in areas of \textit{suspension}.

\subsubsection*{Consensus:}
Regarding the results for the \textit{OA} and  the $MF_1$-score achieved after the \textit{consensus} training of the networks, significant improvements of up to 3.6\% are obtained for the R-S-Net while only small differences occur in case of the Unet architecture. 
It is noteworthy, that in this setting the R-S-Net achieves a better OA and $MF_1$-score than the Unet. 
The class-wise values for \textit{recall} and \textit{precision} allow for deeper insights into the effect caused by the consensus training using the additional unlabelled training data. 
While the \textit{precision} of the minority class \textit{aggregate} increases significantly by 12.3\% and 13.9\% for the Unet and the R-S-Net, respectively, the \textit{recall} of that class decreases by 12.4\% and 6.7\%. In contrast, the effect for the majority class \textit{suspension}, reveals an opposite behaviour, i.e.\ the consideration of the consensus loss leads to an enhancement of the \textit{recall} but to a decrease of the \textit{precision} results, although the magnitude of the differences is smaller compared to the ones of the class \textit{aggregate}. 
We consider these effects being directly related to the \textit{blind spot} of the consensus principle described in Sec.~\ref{sec:BlindSpot}: Because the same incorrect prediction of both, the \textit{main} and the \textit{auxiliary} branches, are not penalised by the consensus loss $\mathcal{L}_\text{cons}$ and at the same time, those cases are more likely to occur for more frequent classes (\textit{suspension} in this case), the training of the segmentation networks following the consensus principle favours the prediction of majority class labels. 
As a consequence, the absolute number of predicted labels belonging to the majority class is likely to increase while the number of minority class labels tends to decrease, causing the recall of the majority class to become larger and the recall of the minority class to become smaller, as observable from Tab.~\ref{tab:Results}.
This effect is also clearly visible in Fig.~\ref{fig:QualRes}.
Comparing the qualitative results obtained by the \textit{base} and the \textit{cons} variant, a distinct decrease of the FP \textit{aggregate} classifications (red areas) can be seen, while the amount of FN segments (blue areas) increases. The latter effect mostly leads to the misclassification of complete aggregate particles by the \textit{cons} setting, which were successfully detected by using the \textit{base} variant.
 
\subsubsection*{Consensus+prior:}
\begin{figure}[H]
\begin{center}
		\includegraphics[width=1.0\columnwidth]{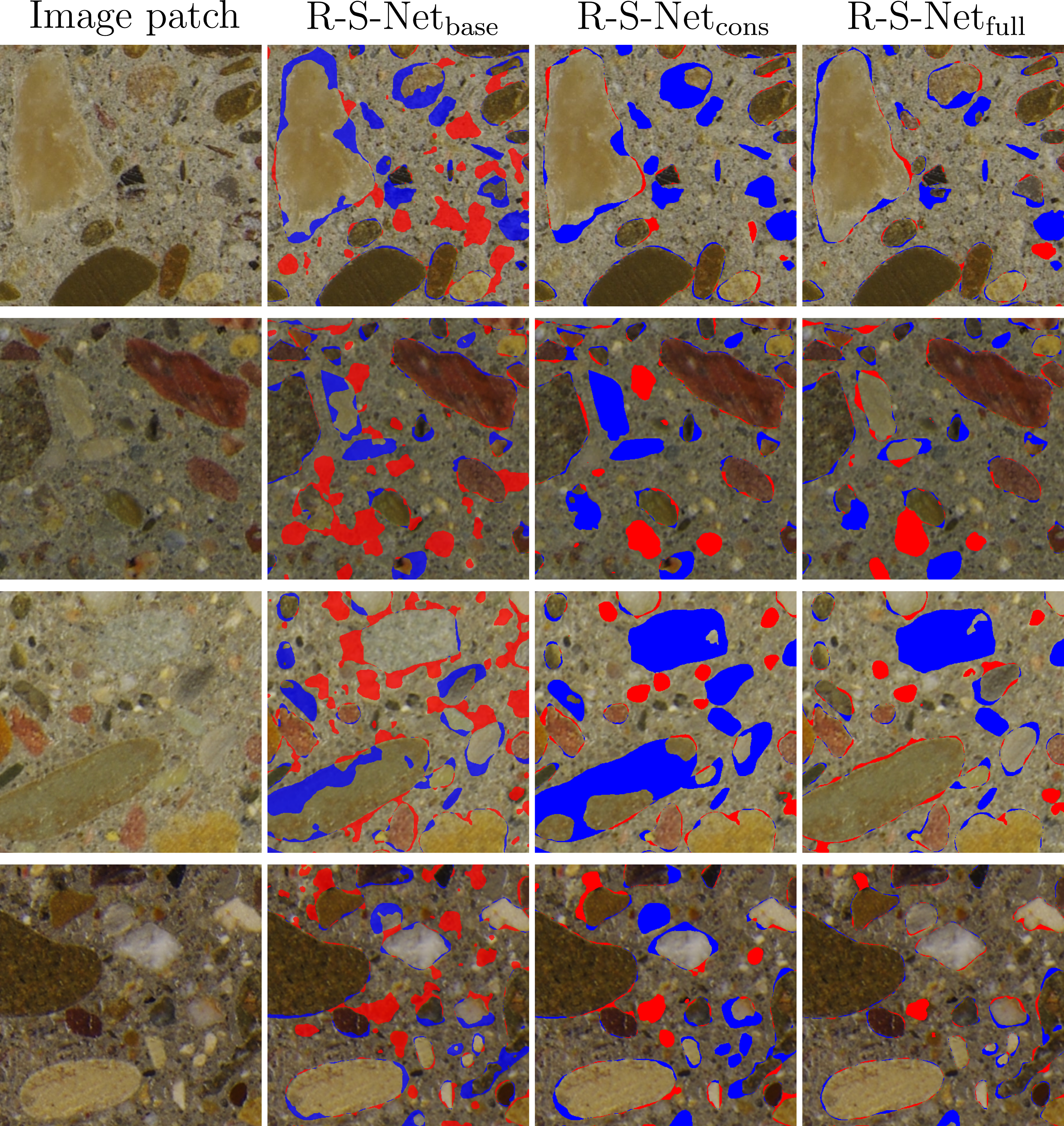}
	\caption{Qualitative results for the different settings of the proposed framework of the R-S-Net achieved for the $T_1$ training setup. Correctly classified pixels are shown without additional colour coding, FN \textit{aggregate} pixels are coloured in \textbf{blue} and FP \textit{aggregate} pixels are coloured in \textbf{red}.}
\label{fig:QualRes}
\end{center}
\end{figure}

The goal of this paper is to propose a strat\-egy to counteract the effect caused by the \textit{blind spot} of the consensus principle by introducing prior information as additional training signal to the semi-supervised segmentation framework.
As can be seen from Tab.~\ref{tab:Results}, considering the \textit{full} framework during training leads to a significant increase of the \textit{recall} values of the minority class \textit{aggregate} by 3.2\% for the R-S-Net and by even 9.8\% in case of the Unet architecture, compared to the consensus solution. In contrast, the values for \textit{precision} of that class decrease, but by a smaller margin. Again, the behaviour of the values for these metrics achieved for the class \textit{suspension} is vice-versa. 
Accordingly, it can be seen from the qualitative results in Fig.~\ref{fig:QualRes}, that the \textit{full} setting of our proposed semi-supervised segmentation framework, distinctly reduces the amount of FN classifications (blue) of the class \textit{aggregate}, while the effect on the FP classifications (red) are only marginal.
Finally, for both architectures, the consideration of the proposed prior losses during training in the \textit{full} framework leads to the best values for the $F_1$-score as well as for \textit{OA} and $MF_1$, proving the suitability of the proposed additional regularisations for semi-supervised consistency training. 
 
%*******************************
%*******************************

%==================================================================================
%
\begin{figure*}[ht] 
\centering
%\hspace{-1.0cm}
\subfloat[Overall accuracy] {\includegraphics[width=0.33\textwidth]{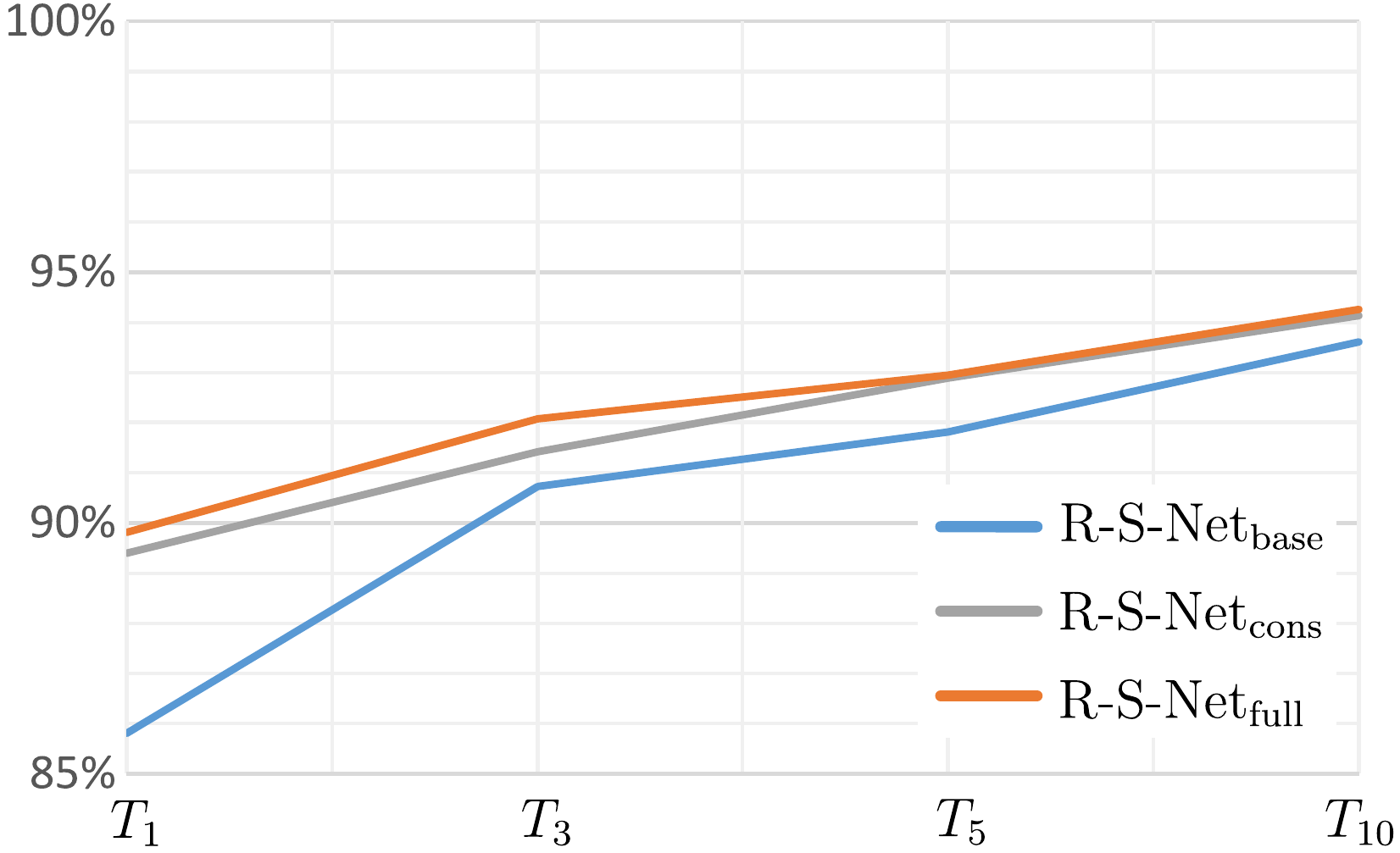}\label{fig:ResOA}} 
\hspace{0.01cm}
\subfloat[$F_1$-score for the class \textit{aggregate}.] {\includegraphics[width=0.33\textwidth]{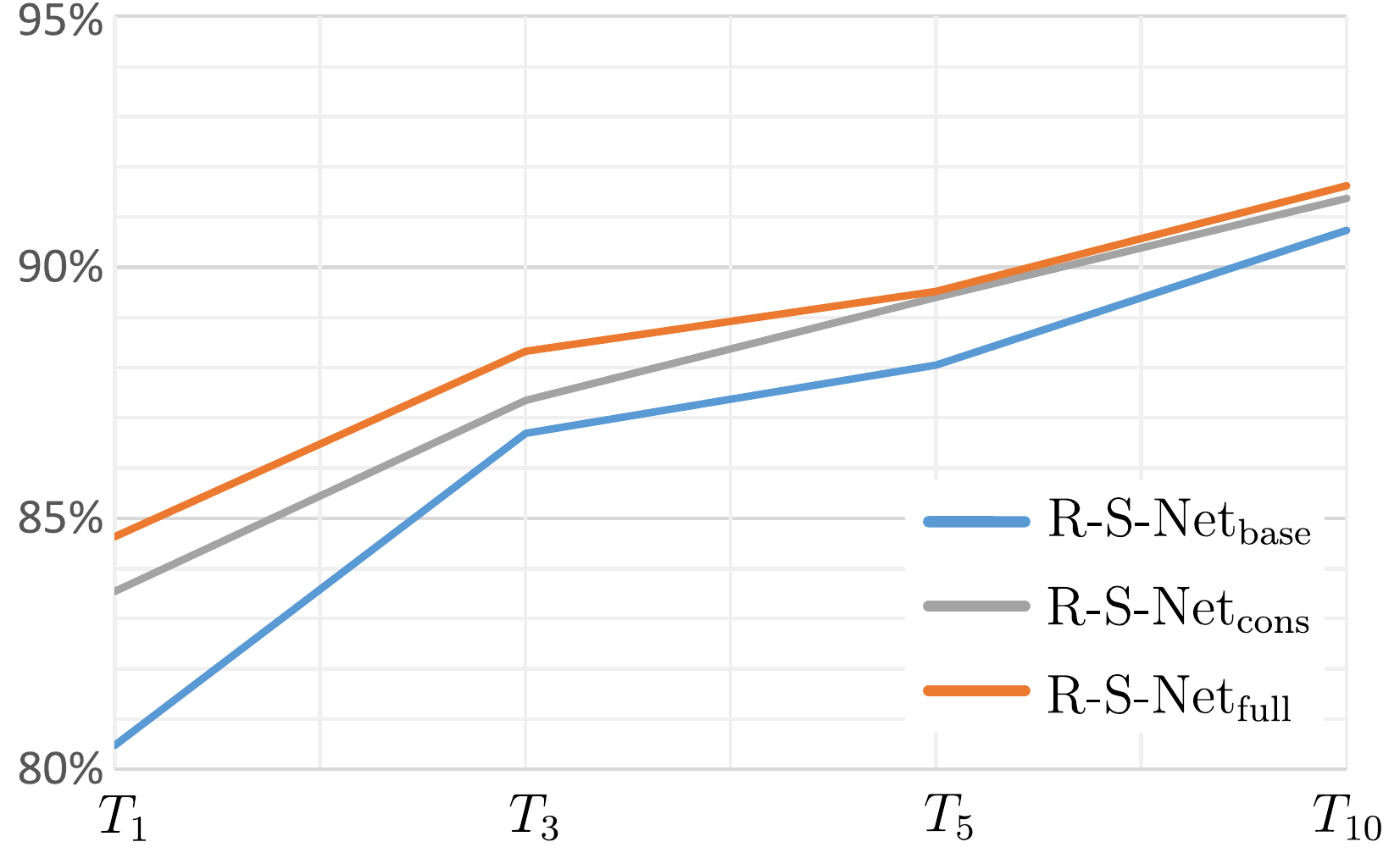}\label{fig:ResAgg}} 
\subfloat[$F_1$-score for the class \textit{suspension}.] {\includegraphics[width=0.33\textwidth]{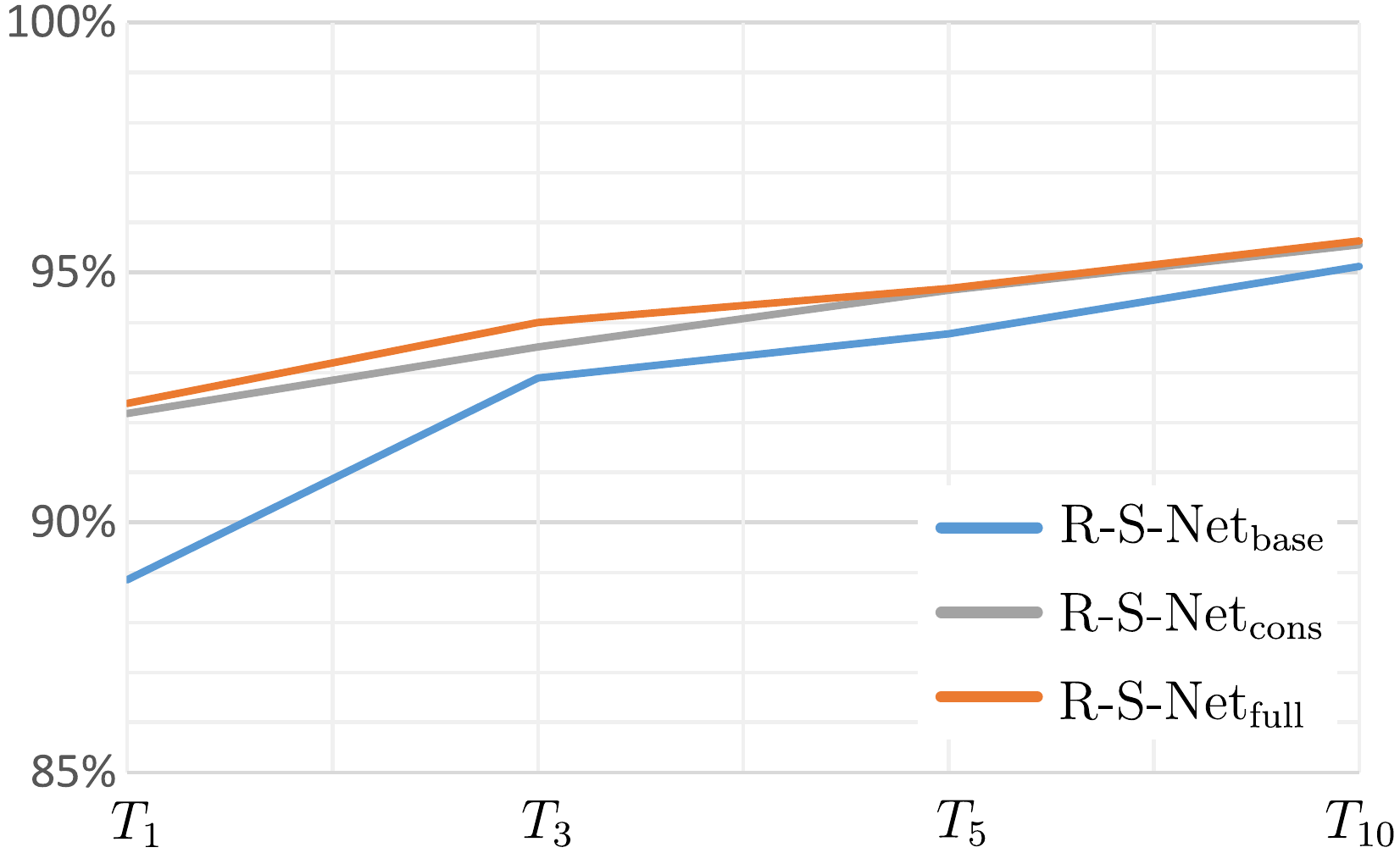}\label{fig:ResSusp}} 
\caption{Results of the ablation study on the effect of different amounts of labelled data used for training.}
\label{fig:Results} 
\end{figure*}
%
%==================================================================================

In Fig.~\ref{fig:Results} we show the results for OA and the class-wise $F_1$-scores of our ablation study on the effect of the amount of labelled data ($T_1$ - $T_{10}$) considered during training on the example of the R-S-Net architecture and for the three investigated framework variants \textit{base}, \textit{cons}, and \textit{full}.
As can be seen, increasing the amount of labelled data for training also increases the performance of all three variants.
In this context, the largest improvements are achieved between the setups $T_1$ and $T_3$, between which the amount of training data is tripled. 
Here, the OA increases by 4.9, 2.1, and 2.2\% for the \textit{base}, \textit{cons}, and \textit{full} variants, respectively. 
Between the training setups $T_3$ and $T_{10}$, further enhancements of 2.9, 2.7, and 2.5\% for the OA are achieved by the different variants.\\
Inspecting the $F_1$-scores obtained for both classes, it is apparent that, while both classes profit from the consideration of more labelled data during training, the effect for the minority class \textit{aggregate} is larger compared to the one for the class \textit{suspension}.
While the $F_1$ score of the class \textit{aggregate} increases by up to 10.3\% between the $T_1$ and $T_{10}$ training variants, the enhancement for the class \textit{suspension} is distinctly smaller, namely only 6.3\%.
\sloppy
Furthermore, it can be seen that the effect of using unlabelled data for the semi-super\-vised segmentation learning on the quality measures for both, OA and $F_1$-scores, is largest in the case of very few annotated training data ($T_1$), while the differences between the results of the purely supervised and the semi-super\-vised variants decrease the more labelled data is available for training. 
Still, our proposed approach achieves the biggest enhancement of OA and $F_1$-scores of both classes in the case where only few annotated training data is considered and achieves the best results for the quality measures among all settings considered in Fig.\ref{fig:Results}.

\sloppy % KAO: Sloppy spacing ensures non-overfull lines. Can be removed if this is not an issue.

%========================================================================================================
\section{Conclusion} \label{sec:Conclusion}
In this paper, we present a novel framework for semi-supervised semantic segmentation based on consensus training. We identify limitations inherent to the consensus principle and propose additional regularisation techniques based on prior knowledge about the class distribution and on auto-encoder constraints to overcome these limitations. 
We demonstrate superior results achieved by our proposed strategy compared to purely supervised and standard semi-supervised training and present a new light-weight architecture achieving competing results to a state-of-the-art heavy-weight architecture on our new concrete aggregate data set.
In the future, we aim at a more in-depth analysis on the influence of the individual prior losses and their weights, additional variations of perturbation functions, and the consideration of multiple auxiliary branches in the framework in order to investigate the effect of the individual components on the semi-supervised training behaviour. Also, we want to apply the proposed framework on multi-class segmentation tasks. 
Besides, we want to make use of the segmentation results to derive information about the segregation behaviour and stability properties of the concrete. To this end, we will develop methods for an automatic inference of relevant evaluation criteria as e.g.\ the sedimentation limit and the grain size distribution from the segmentations.

\sloppy % KAO: Sloppy spacing ensures non-overfull lines. Can be removed if this is not an issue.

\renewcommand{\bibsection}{\section*{REFERENCES}}
{
	\begin{spacing}{1.1}
		\normalsize
		\bibliography{../../Literatur} % Include your own bibliography (*.bib), style is given in isprs.cls
	\end{spacing}
}

\end{document}